\documentclass[conference,10pt]{IEEEtran}
\IEEEoverridecommandlockouts
\usepackage{silence}
\usepackage{cite}
\usepackage{graphicx}
\usepackage{textcomp}
\usepackage{xcolor}
\def\BibTeX{{\rm B\kern-.05em{\sc i\kern-.025em b}\kern-.08em
    T\kern-.1667em\lower.7ex\hbox{E}\kern-.125emX}}

\usepackage{amssymb}
\usepackage{amsthm}
\usepackage{comment}
\theoremstyle{plain}

\theoremstyle{definition}

\usepackage[fleqn]{mathtools}

\usepackage{bussproofs}
\usepackage{proof}
\usepackage{tabularx}
\usepackage{booktabs}
\usepackage{array}
\usepackage{float}
\usepackage{caption}
\usepackage{graphics}
\usepackage{subfig}
\usepackage{setspace} 
\usepackage{framed}
\usepackage{algorithm}
\usepackage{graphicx,wrapfig,lipsum}
\usepackage{algpseudocode}
\usepackage[utf8]{inputenc} 
\usepackage[T1]{fontenc}   
\usepackage{nicefrac}       
\usepackage{microtype}
\usepackage{hyperref}
\usepackage{enumitem}
\usepackage{multirow}
\usepackage{xcolor} 
\usepackage{amssymb}
\usepackage{amsmath}
\usepackage{amsfonts}
\usepackage{amsthm}
\usepackage{balance}
\theoremstyle{definition} \newtheorem{theorem}{Theorem}[section]
\theoremstyle{definition} \newtheorem{definition}[theorem]{Definition}
\theoremstyle{definition} 
\newtheorem{proposition}{Proposition} 
\newtheorem{axiomset}{Axiom Set}
\begin{document}
\title{Deontic Temporal Logic for \\ Formal Verification of AI Ethics\\}
\author{
    Priya T.V., \quad Shrisha Rao \\
    International Institute of Information Technology - Bangalore \\
    \texttt{priya.tv@iiitb.ac.in, shrao@ieee.org}
}

\maketitle
\begin{abstract}
 Ensuring ethical behavior in Artificial Intelligence (AI) systems amidst their increasing ubiquity and influence is a major concern the world over.  The use of formal methods in AI ethics is a possible crucial approach for specifying and verifying the ethical behavior of AI systems. This paper proposes a formalization based on deontic logic to define and evaluate the ethical behavior of AI systems, focusing on system-level specifications, contributing to this important goal. It introduces axioms and theorems to capture ethical requirements related to fairness and explainability. The formalization incorporates temporal operators to reason about the ethical behavior of AI systems over time. The authors evaluate the effectiveness of this formalization by assessing the ethics of the real-world COMPAS (recidivism risk prediction) and loan (approval) prediction AI systems.  Various ethical properties of COMPAS and loan prediction systems are formalized using temporal and deontic logic, enabling an automated theorem prover to verify whether these systems satisfy the specified properties. The formal verification reveals that both systems fail to fulfill certain key ethical properties related to fairness and non-discrimination, while satisfying others. This demonstrates the effectiveness of the proposed formalization in identifying potential ethical issues in real-world AI applications.
\end{abstract}

\begin{IEEEkeywords}
Artificial Intelligence, Ethics, Deontic Temporal Logic
\end{IEEEkeywords}

\section{Introduction} \label{introduction}
Artificial Intelligence (AI) systems are becoming increasingly ubiquitous and influential in our lives, making decisions that can have significant ethical implications. As AI continues to advance and take on more complex tasks, it is crucial to ensure that these systems behave ethically 
 \cite{r9_global, r_10surveyonethics, r11_whymachineethics, r12_machineethics, r13_aiethics, r14_aiethicsguidelines,needofaiethics,needofaiethics1,needofaiethics2}. However, defining and enforcing ethical behavior in AI is a challenging task, as ethics often involve abstract concepts and context-dependent judgments\cite{r18_ethicsandprivacy,r19_ethicsissuesinitiative,r29_ethicsoverview}. There are numerous principles generated by various organizations and regulation bodies. For instance, the Ethically Aligned Design (EAD) guidelines of IEEE recommend that AI design prioritize maximizing benefits to humanity~\cite{r3_ieee_initiative}. Furthermore, The European Commission has released Ethics Guidelines for Trustworthy AI, stressing the importance of AI being human-centric~\cite{EUguidelines}. The national plan for AI in the United Kingdom suggests the establishment of an AI Code~\cite{r4_aicodeuk}. Australia has also introduced its AI ethics framework~\cite{r5_australiafm}, which adopts a case study approach to examine fundamental ethical principles for AI and offers a toolkit for integrating ethical considerations into AI development. Adding to this are Beijing's AI principles, Amnesty International ACM code of ethics, and many more. In addition to governmental organizations, prominent companies such as Google~\cite{r6_google} and SAP~\cite{r7_sap2021saps} have publicly released their AI principles and guidelines. Moreover, professional associations and non-profit organizations like the Association for Computing Machinery (ACM) have issued their recommendations for ethical and responsible AI~\cite{ACM_everaging,ACMGUIDELINES}.  
Despite these efforts, a consensus on the ethics of AI remains challenging.  They lack a unified framework of guidelines that can be universally adopted by organizations, governments, and regulatory bodies to formulate and assess the ethics of systems. It is not yet clear what common principles and values AI should adhere to. Establishing cohesive and widely accepted ethical principles for AI is crucial across different organizations and domains. Moreover, ethics is a philosophical question of what is right or wrong~\cite{r15_philosophy,r16_philosophy}. Its qualitative nature makes it complex and hard to define precisely, and hence needs a \textit{mathematically rigorous framework}. \footnote{While a mathematically rigorous framework can assist in clarifying and reasoning about ethical principles, particularly in system-level verification, we recognize that such formalization may abstract away elements of moral deliberation, human judgment, and situational complexity~\cite{formalization_limitation}. Our use of mathematical formalism is therefore intended as a tool to support—rather than replace—the richer, dialectical exploration that ethical reasoning necessitates.}

\par
To address this challenge, we are exploring the use of formal methods to express and prove the ethical correctness of AI systems. One promising approach is the use of \textit{deontic logic}, a branch of modal logic that deals with concepts such as obligation, permission, and prohibition~\cite{r17_deontologicalethics,r_34applicationsdeonticlogic}. Deontic logic provides a rigorous framework for reasoning about ethical norms and can be used to formalize ethical principles~\cite{r_30aiprinciples} and constraints. Several works have explored using deontic logic to formalize machine ethics, mainly for robots~\cite{bringsjord2006toward,ethicalrobots} and normative systems~\cite{boella2006introduction}. These studies have concentrated on Kantian ethics, integrating deontic and temporal logic to verify the ethical behavior of autonomous systems, such as unmanned aircraft, over time~\cite{r_25kantian,dennis2016formal}. 
\par
While promising, these methods are often constrained by specific ethical frameworks and fail to scale with the complexity of modern AI, which increasingly mimics human tasks, leverages natural language processing, and operates on vast datasets. This leads to a proliferation of potentially subjective ethical rules influenced by personal biases. The dynamic, evolving nature of AI further complicates ethical formalizations~\cite{seshia2022toward}. Critically, many of these approaches remain theoretical, lacking practical integration with machine learning techniques, highlighting the need for more adaptive and implementable ethical frameworks in AI~\cite{Logic-BasedEthicalPlanning}. Our work represents a foundational effort to develop a unified framework that addresses ethical principles in AI systems, with a specific focus on granular levels of explainability and fairness. It builds upon existing approaches~\cite{dlforethics,bringsjord2006toward} in specialized domains, extending them to tackle the unique challenges posed by modern AI ethics. In doing so, this paper introduces a novel direction for formalizing and verifying the ethical principles of AI systems. While the proposed framework is broadly applicable to the ethics verification of autonomous systems, this work specifically focuses on its application to AI. The scope of this work is to provide a conceptual foundation and framework for formalizing AI ethics using deontic and temporal logic. Rather than focusing on individual actions or decisions, our approach emphasizes system-level specifications. It involves defining ethical properties that AI systems should ideally meet. For example, an AI system that uses gender as a feature should avoid making decisions explicitly based on it. These properties ensure that systems are designed to identify and mitigate biases and ethical violations effectively. Defining properties like "forbidden to consider sensitive features in predictions" provides a way to analyze commonly discussed properties of AI in a unified manner. This abstraction also helps to reduce the difficulty in formalizing each action or each type of individual~\cite {seshia2022toward}. Additional details on formalizing the properties of AI systems and their verification are provided in Section~\ref{implementation}. We demonstrate the feasibility of our framework through the formal verification of properties in a classification-based setting. However, in principle, the framework can be extended to a wide range of AI paradigms, including large language models (LLMs), reinforcement learning (RL), and beyond.

\par
The basic model for applying deontic logic to AI ethics uses first-order logic to define predicates (Table~\ref{table-predicates}) and axioms that capture ethical requirements. This model introduces variables such as $\mathit{x}$ to indicate an AI system, $\mathit{x}_a$ to indicate that system $\mathit{x}$ performs an action $\mathit{a}$, and predicates such as $\mathcal{E}(x)$ to indicate ethical behavior. Axioms~\ref{axiom1}--~\ref{axiom1.5} are then defined using these predicates to express ethical obligations, prohibitions, and permissions for AI systems. Building upon the basic model, an extended version incorporates temporal operators from temporal logic to reason about the ethical behavior of AI systems over time. This extension allows for the expression of more complex ethical requirements, such as the obligation for AI systems to maintain fairness over time or the prohibition of exhibiting bias. The temporal operators used in this model  include ``always'' ($\Box$), ``eventually'' ($\diamond$), and ``until'' ($\mathcal{U}$) presented by Manna and Pnueli~\cite{r_35temporallogic}.  In this framework, $\mathit{x}$ persists across states, while actions cause state transitions. If no action influences the transition, the passage of time follows a default evolution independent of AI choices.
\par
Theorems~\ref{theorem1} and~\ref{theorem2} in the basic model of deontic logic for AI ethics explore the relationships between ethical obligations, prohibitions, and permissions for AI systems. These theorems employ first-order logic and the defined predicates to derive conclusions about the ethical behavior of AI systems. The proofs of these theorems rely on techniques such as modus ponens, contraposition, and proof by contradiction to establish the logical connections between the axioms and the derived statements. The general flavor of the theorems is to provide a rigorous foundation for reasoning about the ethical requirements of AI systems, demonstrating how the Axioms~\ref{axiom1}--\ref{axiom1.5} can be used to infer specific obligations, prohibitions, and permissions in various contexts. By establishing these logical relationships, the theorems contribute to a comprehensive framework for analyzing and ensuring the ethical behavior of AI systems. Similar to this, the Theorems~\ref{theorem3} to~\ref{theorem8} in the extended model, which incorporates temporal logic operators, explore the ethical behavior of AI systems over time. These theorems focus on capturing the temporal aspects of ethical requirements, such as the obligation to maintain fairness or the prohibition of exhibiting bias. The proofs of these theorems utilize the semantics of the temporal operators, such as $\Box$(always), $\diamond$(eventually), and $\mathcal{U}$(until), in conjunction with the Axiom lists~\ref{A2} to~\ref{A3} and predicates defined (Table~\ref{table-predicates}) in the basic model. The general flavor of the theorems in the extended model is to provide a more expressive and nuanced framework for reasoning about the ethical behavior of AI systems, considering the dynamic and evolving nature of these systems. The theorems establish logical connections between the temporal properties of AI systems and their ethical obligations, allowing for the analysis of more complex and realistic scenarios. By incorporating temporal aspects, the extended model enables a deeper understanding of the long-term ethical implications of AI systems and provides a foundation for designing and verifying AI systems that behave ethically over time.
\par
The importance of this work lies in formalizing the axioms to define the ethical requirements of an AI system and its potential to provide a formal and verifiable framework for ensuring the ethical behavior of AI systems. Our experimental findings show the effectiveness of this formalization in assessing the ethics of real-world AI systems---Loan prediction and COMPAS. We evaluated the ethical aspects of the systems, wherein we defined specific properties that these systems must adhere to to be deemed ethical. Our results revealed that certain properties were indeed satisfied by the system, while others were not (Table~\ref{table-result}). The results demonstrated that applying deontic logic and temporal operators to AI ethics represents a significant step forward in formally specifying and verifying the ethical behavior of AI systems. 
\par
Section~\ref{section2} discusses the related works. Section~\ref{deontic logic for reasoning} introduces temporal deontic logic for AI ethics formalization. Subsections~\ref{formalizing fairness} and~\ref{formalizing explainability} address fairness and explainability principles. Section~\ref{implementation} covers the application of this approach to real-world AI systems, including algorithms~\ref{algm1} and~\ref{algm2} to demonstrate the implementation of this method on real datasets, providing readers with detailed insights into how it is executed. Section~\ref{conclusion} concludes.
\section{Related Works} \label{section2}
The field of ethical reasoning encompasses a range of approaches, often grounded in formal logic, to ensure trustworthy and morally sound behavior in autonomous systems. Several works contribute to this domain, presenting unique methodologies and frameworks to address the complex interplay between ethics and machine decision-making. Among these, deontological ethics, particularly Kantian frameworks, are well-suited for machine ethics due to their rule-based nature. This method ensures that machines refrain from harmful actions through rule-based formalization~\cite{deontologicalmachineethics,AutomatedReasoningDeonticLogic,dlforethics}.
\par
The earlier work introduces the GenEth ethical dilemma analyzer~\cite{GenEth}, which utilizes inductive logic programming to infer principles for ethical actions. Dominance Act Utilitarianism (DAU), a deontic logic of agency, is another framework for encoding and analyzing obligations in autonomous systems. DAU frameworks are efficient in addressing safety-critical behaviors, such as adherence to traffic laws and avoidance of reckless actions~\cite{deonticlogicanalysisofautonomoussystems'safety}. Such frameworks can formalize ethical obligations in systems like self-driving cars, enabling systematic reasoning about social and moral responsibilities~\cite{AlgorithmicEthicsFormalizationandVerificationofAutonomousVehicleObligations}. Additionally, several works employ the Belief-Desire-Intention (BDI) framework to formalize reasoning about moral agents~\cite{LOGICFORREASONINGABOUTMORALAGENTS,OnProactiveTransparentandVerifiableEthicalReasoningforRobots,dennis2016formal}. This structure supports transparency and formal verification in ethical decision-making processes for robots.
\par
Further, the literature explores the use of high-level action languages and Answer Set Programming to design ethical autonomous agents~\cite{DeclarativeModularFrameworkforRepresentingandApplyingEthicalPrinciples,Ethicalsystemformalizationusingnon-monotoniclogics,deonticLogicforProgrammingRightfulMachines}. There are several works that propose using deontic logic to constrain robot behavior in ethically sensitive environments, as this type of logic helps interpret natural language directly~\cite{TowardsVerifiablyEthicalRobotBehaviour,bringsjord2006toward,ethicalrobots}. These frameworks are also used for ethical reasoning in the healthcare domain, emphasizing accountability and transparency~\cite{EthicsinDigitalHealth:adeonticaccountabilityframework}. Additionally, various works focus on using deontic logic-based frameworks for formalizing ethical reasoning in AI systems~\cite{DeepLearningOpacityandtheEthicalAccountabilityofAISystems,DeontologyandSafeArtificialIntelligence,Formalethicalreasoninganddilemmaidentificationinahuman-artificialagentsystem}.
\par
To accommodate the dynamic nature of machine environments~\cite{Ensuringtrustworthyandethicalbehaviourinintelligentlogicalagents}, several studies propose integrating deontic logic with temporal operators, facilitating the representation of concepts like refraining from specific actions or opting for alternative actions~\cite{temporaldynamicdeonticlogic,DeonticLogicReasoningInfrastructure}. This extension facilitates a richer understanding of ethical constraints in dynamic environments. Furthermore, frameworks combining linear temporal logic with lexicographic preference modeling support ethical decision-making in robotics~\cite{Logic-BasedEthicalPlanning}.
\par
Thus, the literature suggests that rule-based ethical theories, particularly deontology, are essential for developing trustworthy AI systems~\cite{r_25kantian}. However, considering the dynamic nature of AI, especially regarding fairness and explainability at a granular level, significant gaps remain. Our work addresses these gaps by introducing fairness and explainability at multiple granularities, including stable, transient, inherent, and retrofitted/artificial dimensions. These distinctions capture the evolving nature of AI and its complex decision-making processes, providing a more comprehensive approach to ethical verification. Furthermore, existing frameworks often overlook the impact of personal biases introduced during training and lack mechanisms to mitigate them effectively. To address this, we propose an iterative learning approach designed to identify and reduce the influence of personal biases in the system. While prior research highlights the gap between theoretical ethical reasoning and its practical application in autonomous agents~\cite{Logic-BasedEthicalPlanning}, our framework bridges this divide. By implementing and testing the framework in real-world AI systems such as COMPAS and Loan prediction systems, we validate its effectiveness and ensure its applicability. 
\par
A key feature of our approach is the generation of counterexamples that illustrate how specific properties may violate system specifications. This not only strengthens the verification process but also provides actionable insights for refining system behavior. Furthermore, we leverage theorem provers to capture and validate properties derived from real-world data distributions and predictions, ensuring alignment with ethical principles under varying conditions.  The literature survey presented in this paper provides a focused overview to contextualize the study, acknowledging the possibility of additional relevant works in the field.

\section{Deontic Logic for Ethics} \label{deontic logic for reasoning}

\subsection{Preliminaries} \label{deontic logic}
 Deontic logic is a branch of symbolic logic that deals with normative concepts such as obligation ($O$), permission ($P$), and forbidden ($\neg P$). Our work provides the reader with insight into the use of Deontic Logic to formalize and verify the ethical principles of an AI system. The principles that we focus in this work include fairness and explainability. The ethics of AI is more a philosophical question about what is morally right or wrong, permissible or impermissible. By representing ethical rules as deontic statements, AI designers can specify what a system ought or ought not to do. They can evaluate actions or decisions against a set of predefined ethical rules and determine whether the system complies with these rules. This is essential to guarantee that AI systems act morally following societal norms. 
 \par
Standard Deontic Logic (SDL) represents a foundational framework within deontic logic. In this work, we build upon SDL by extending it with temporal operators to enable reasoning about ethical properties over time~\cite{temporaldynamicdeonticlogic,deonticLogicforProgrammingRightfulMachines,TDLSEMANTICS}.  
SDL formulas include classical propositional logic and it operates as a monadic deontic logic, meaning its operators (obligation, permission, forbidden) apply to individual formulas ($\varphi$); they are read as "it is obligatory that $\varphi$", "it is permissible that $\varphi$", and "it is forbidden that $\varphi$" respectively. Furthermore, they are cross-definable. For instance, $P\varphi:=\neg O(\neg \varphi)$, and $F\varphi:=O \neg \varphi$ . This logical statement explains that permission ($P$) or forbidden ($\neg P$) can be represented in terms of obligation ($O$).
 Temporal Deontic Logic (TDL) expands SDL by integrating temporal aspects into norms and obligations, introducing operators such as $always (\Box), eventually (\diamond), next$, and $ until (\mathcal{U})$. For instance, $next \ \varphi$ means that the proposition $\varphi$ holds in the next time step. Similarly, $\varphi \mathcal{U} \psi$ means $\varphi$ is true until $\psi$ becomes true. 
 We use the semantics of the combined logic as an extension of the Kripke-style possible world semantics of deontic logic with temporal operators, as described in reference~\cite{TDLSEMANTICS}. While branching-time logic is often used to reflect future uncertainty, we adopt Linear Temporal Logic (LTL) due to its simplicity and relevance to AI verification, where obligations typically unfold along a single execution path. Moreover, LTL allows us to capture obligations or constraints that must persist, eventually hold, or be updated dynamically as the system learns or acts in a sequence of decisions. This helps to preserve sequential consistency rather than enabling arbitrary time jumps. We recommend that interested readers refer to \cite{handbook_deontic,TDLSEMANTICS} for further details on the foundational principles. 
\par
In this section, we focus on formalizing the overall ethical behavior of an AI system. For the formalization, we use the predicates as shown in Table~\ref{table-predicates}. The predicate is a function that takes an input and returns a truth value. 
\begin{definition}[TDL syntax] \label{defn_tdlsyntax}
    Given a set $p$ of atomic propositions, the temporal deontic logic, TDL is defined as,\\
    $\varphi ::= p \mid \neg \varphi \mid \varphi \lor \psi \mid \varphi \land \psi \mid \varphi \to \varphi \mid O\varphi \mid P\varphi \mid \neg P\varphi \mid \Box\varphi \mid \diamond\varphi \mid \varphi \mathcal{U} \psi \mid \forall v. \varphi \mid \exists v. \varphi$\\
    
\end{definition}

We define the semantics of Temporal Deontic Logic (TDL) based on the foundational concepts presented in the work by \cite{tdl_semantics,TDLSEMANTICS}.

\begin{definition}[TDL Semantics]\label{defn-TDLSemantics}

Let $\mathcal{A}$ be a set of atomic propositions.
A TDL model is a structure
\[
\mathfrak{M} = (W, \mathbb{T}, \{R_D^t\}_{t \in \mathbb{T}}, D, V)
\]
where:
\begin{itemize}
    \item \ $W$ is a non-empty set of worlds;
    \item \ $\mathbb{T} = \mathbb{N}$ is an infinite, linearly ordered set of time points;
    \item \ for each $t \in \mathbb{T}$, $R_D^t \subseteq W \times W$ is a serial deontic accessibility relation;
    \item \ $D$ is a constant domain of individuals;
    \item \ $V : \mathbb{T} \times W \rightarrow 2^{\mathcal{A}}$ is a valuation function.
\end{itemize}

For $t \in \mathbb{T}$ and $w \in W$, $V(t,w)$ is the set of atomic propositions true at world $w$ at time $t$.

Let $\eta : \mathrm{Var} \rightarrow D$ be a variable assignment.
Truth of a formula $\varphi$ at time $t$ and world $w$ is written:
\[
\mathfrak{M}, t, w, \eta \models \varphi.
\]
The satisfaction relation is defined inductively as follows:
\begin{flushleft}
\begin{tabular}{ll}
$\mathfrak{M}, t, w, \eta \models p$
& $\iff p \in V(t,w)$ \\

$\mathfrak{M}, t, w, \eta \models \neg \varphi$
& $\iff \mathfrak{M}, t, w, \eta \not\models \varphi$ \\

$\mathfrak{M}, t, w, \eta \models \varphi \land \psi$
& $\iff \mathfrak{M}, t, w, \eta \models \varphi$ \\ & and $\mathfrak{M}, t, w, \eta \models \psi$ \\

$\mathfrak{M}, t, w, \eta \models \varphi \lor \psi$
& $\iff \mathfrak{M}, t, w, \eta \models \varphi$ \\ & or $\mathfrak{M}, t, w, \eta \models \psi$ \\

$\mathfrak{M}, t, w, \eta \models \varphi \rightarrow \psi$
& $\iff \mathfrak{M}, t, w, \eta \models \varphi$ \\ &implies $\mathfrak{M}, t, w, \eta \models \psi$ \\

$\mathfrak{M}, t, w, \eta \models X \varphi$
& $\iff \mathfrak{M}, t+1, w, \eta \models \varphi$ \\

$\mathfrak{M}, t, w, \eta \models \diamond \varphi$
& $\iff \exists u \ge t\ \text{s.t.}\ \mathfrak{M}, u, w, \eta \models \varphi$ \\

$\mathfrak{M}, t, w, \eta \models \Box \varphi$
& $\iff \forall u \ge t,\ \mathfrak{M}, u, w, \eta \models \varphi$ \\

$\mathfrak{M}, t, w, \eta \models \varphi \mathcal{U} \psi$
& $\iff \exists u \ge t$ such that \\
& \quad $\mathfrak{M}, u, w, \eta \models \psi$ and \\
& \quad $\forall v$ with $t \le v < u,\ \mathfrak{M}, v, w, \eta \models \varphi$ \\

$\mathfrak{M}, t, w, \eta \models O \varphi$
& $\iff \forall w' \in W,\ (w, w') \in R_D^t$, \\ & $\mathfrak{M}, t, w', \eta \models \varphi$ \\

$\mathfrak{M}, t, w, \eta \models P \varphi$
& $\iff \exists w' \in W,\ (w, w') \in R_D^t $, \\ & $ \mathfrak{M}, t, w', \eta \models \varphi$ \\

$\mathfrak{M}, t, w, \eta \models \forall x.\varphi$
& $\iff \forall d \in D,\ \mathfrak{M}, t, w, \eta[x \mapsto d] \models \varphi$ \\

$\mathfrak{M}, t, w, \eta \models \exists x.\varphi$
& $\iff \exists d \in D,\ \mathfrak{M}, t, w, \eta[x \mapsto d] \models \varphi$
\end{tabular}
\end{flushleft}
\end{definition}

The following set of axioms has been formulated to specify a set of formal conditions under which an AI system may satisfy a particular interpretation of ethical behavior within our chosen framework. When modeling AI systems that require quantification over agents, actions, or time points, it is necessary to extend deontic and temporal logic with first-order logic formulas and quantifiers~\cite{tdf_fol}. This framework serves as a foundational starting point for developing such an extended logical approach in this direction. 

In this work, we demonstrate the proposed logic through its application to a classification-based AI model. The model is formally defined as a function mapping an input feature vector to a discrete set of class labels. This definition aligns with standard formulations commonly used in the machine learning literature~\cite{classification_defn}.

\begin{definition}[Classification Model]

A classification model \( x \) is a function that maps an input feature vector \(  \vec{v} \in \mathbb{R}^d \) to a class label $c_i$ from the set of class labels \( C = \{ c_1, c_2, \dots, c_k \} \). Formally,
\[
x: \mathbb{R}^d \rightarrow c_i
\]
where \( d \) is the dimension of the feature space, and \( k \) is the number of classes.
\end{definition}

\begin{axiomset}{Basic Axioms} \label{A1}

\begin{enumerate}[label=1.\arabic*,leftmargin=2em]
    
   \item \ \textit{If an AI system \( \mathit{x} \) is ethical, then the action \( \mathit{a} \) performed by \( \mathit{x} \) is ethical: \( \forall x \forall a  \ ( \mathcal{E}(x) \rightarrow (\mathit{x_a} \rightarrow \mathcal{E}_{act}(a)) )\).}  \label{axiom1}
    \item \ \textit{An ethical AI system $x$ is forbidden to perform an unethical action $\mathit{a}$:  $ \forall x \forall a  \ ( \mathcal{E}(x) \rightarrow \neg P(\mathit{x_a} \land \neg \mathcal{E}_{act}(a)))$.}\label{axiom2}
    \item \ \textit{For an ethical AI system $x$, performing an ethical action $\mathit{a}$ is permitted: $\forall x \forall a  \ (\mathcal{E}(x) \rightarrow P(\mathit{x_a} \land \mathcal{E}_{act}(a))) $.} \label{axiom3}
     \item \ \textit{Every action $\mathit{a}$ performed by an ethical AI system $\mathit{x}$ is consistent with the ethical guidelines.:$\forall a \ \mathit{x_a}  \rightarrow \mathcal{G}(\mathit{x_a})  $} \label{axiom1.4} 
    \item \ \textit{If an action $\mathit{a}$ performed by an AI system $x$ follows ethical guidelines $\mathcal{G}(\mathit{x_a})$, then the action is ethically required: $\forall x \forall a  \ ( \mathit{x_a} \land \mathcal{G}(\mathit{x_a})  \rightarrow \mathcal{E}_{act}(a))$} \label{axiom1.5}

\end{enumerate}

\end{axiomset}
\hfill
\begin{table} 
  \caption{Predicates used in this work}
  \label{table-predicates}
  \centering
  \begin{tabular}
  { >{\raggedright\arraybackslash}p{1.5cm}    >{\raggedright\arraybackslash}p{5.8cm}   }  
    \toprule
     
   Predicate & Explanation  \\
   \midrule
   {$\mathcal{E}(x)$} &  $x$ exhibits ethical behavior\\
    {$\mathcal{G}(x)$} & $x$ follows ethical guidelines  \\
    
    {$\mathcal{F}(x)$} & $x$ exhibits fairness \\
   {$\mathcal{B}(x)$} & $x$ exhibits bias \\
   {$ \mathcal{L}(x)$} & $x$ learns iteratively  \\
   {$ \mathcal{X}(x)$} & $x$ has inherent explainability  \\
    {$ \mathcal{R}(x)$} & $x$ has retrofit explainability \\
   {$ \mathcal{C}(x,c)$} & $x$ is counterfactually fair given constraint $c$\\
   {$\mathcal{T}(x)$} & $x$ exhibits transparency  \\
   {$\mathcal{E}_{act}(a)$} & An action $a$ is ethically required  \\ 
   {$\mathcal{F}(\mathit{x})_{train}$} & $x$ exhibits fairness in the training sample  \\
   {$\mathcal{F}(\mathit{x})_{deploy}$} & $x$ exhibits fairness during deployment\\

    \bottomrule
  \end{tabular}
\end{table}
\footnote{ \( \mathcal{G}(x) \) represents compliance with formal ethical guidelines, while \( \mathcal{E}(x) \) represents the overall ethical behavior.}

\paragraph{Remark on normative and descriptive axioms.}
\label{rem:normative-descriptive}
The axiom sets in this work contain two kinds of statements.
\emph{Descriptive} (factual) axioms, such as Axiom~\ref{axiom1}, assert material implications about the actual behavior of systems classified as ethical---for instance, that an ethical system performing action~$a$ in fact produces an ethically required outcome.  \emph{Normative} (deontic) axioms, such as Axioms~\ref{axiom2} and~\ref{axiom3}, use the operators~$O$ and~$P$ to express what \emph{ought} or \emph{may} be the case.  In standard deontic logic, the bridge $O(\varphi) \to \varphi$ (``what is obligatory is actual'') does not hold in general.  Our framework avoids this collapse by formulating the core derivations (Theorems~\ref{theorem1} through~\ref{theorem8}) so that factual conclusions follow from factual axioms, while deontic axioms govern permissibility and prohibition.  The descriptive axioms can be understood as characterizing an \emph{idealized compliant} system: one that, by definition, satisfies its ethical obligations.  This modeling choice is analogous to the ``compliant agent'' assumption common in normative multi-agent systems~\cite{boella2006introduction}.

Based on the foundational principles outlined above and the domain knowledge, we have formulated Theorems ~\ref{theorem1} and ~\ref{theorem2} to ensure that the AI system adheres to ethical standards in all relevant dimensions. Theorem \ref{theorem1} formalizes the principle that an AI system adhering to ethical guidelines cannot simultaneously execute an action that is deemed unethical. The theorem expresses a logical safeguard: compliance with ethical guidelines excludes the possibility of unethical behavior. In other words, if a system genuinely follows its prescribed ethical framework, it is logically inconsistent for it to perform an action that violates ethical standards.
\begin{theorem} \label{theorem1}

   If an AI system $x$ follows ethical guidelines, then it cannot perform an action $a$ that is ethically impermissible: $\mathcal{G}(x_a) \rightarrow \neg(\mathit{x_a} \land \neg\mathcal{E}_{act}(a))   $
\end{theorem}

\begin{proof}~
    \begin{enumerate} [leftmargin=2em]
        \item \ Assume $\mathcal{G}(x_a)$.
        \item \ Assume for contradiction that $\mathit{x_a} \land \neg \mathcal{E}_{act}(a)$.
        \item \ From step 2 through conjunction elimination we get,$\mathit{x_a}$ and $\neg \mathcal{E}_{act}(a)$
        \item \ From steps 1 and 3, through conjunction introduction, we have $\mathit{x_a} \land \mathcal{G}(x_a)$.
        \item \ From Axiom~\ref{axiom1.5}: 
       $  \mathit{x_a} \land \mathcal{G}(x_a) \rightarrow \mathcal{E}_{act}(a)
        $ applying modus ponens gives $\mathcal{E}_{act}(a)$.
        \item \ Step 3 gives $\neg \mathcal{E}_{act}(a)$.
        \item \ Contradiction: $\mathcal{E}_{act}(a) \land \neg \mathcal{E}_{act}(a)$.
        \item \ Therefore, the assumption $\mathit{x_a} \land \neg \mathcal{E}_{act}(a)$ is false:$ \neg (\mathit{x_a} \land \neg \mathcal{E}_{act}(a))$
        \item \ Hence, we conclude:
       $ \mathcal{G}(x_a) \rightarrow \neg (\mathit{x_a} \land \neg \mathcal{E}_{act}(a))$
    \end{enumerate}
\end{proof}

Theorem~\ref{theorem2} states that an AI system cannot be in a state where it is required to do something but not allowed to do it. This maintains logical and ethical consistency. Such a condition is fundamental in designing AI systems that reason ethically, as it ensures they are never blocked from doing what is morally required.
\begin{theorem}\label{theorem2}
  An ethical AI system is not permitted to refrain from an ethically required action $\mathit{a}$: $
O(\mathit{x_a} \land \mathcal{E}_{act}(a)) \rightarrow \lnot P(\lnot (\mathit{x_a} \land \mathcal{E}_{act}(a)))
$
  
\end{theorem}
\begin{proof}
Assume $O(\mathit{x_a} \land \mathcal{E}_{act}(a))$ (the action is ethically required). We use the standard deontic axiom: $O(p) \rightarrow \lnot P(\lnot p)$.

Let $p = \mathit{x_a} \land \mathcal{E}_{act}(a)$. Then:

\[
\infer[] 
  {\lnot P(\lnot (\mathit{x_a} \land \mathcal{E}_{act}(a)))}
  {
    \infer[]
      {O(\mathit{x_a} \land \mathcal{E}_{act}(a)) \rightarrow \lnot P(\lnot (\mathit{x_a} \land \mathcal{E}_{act}(a)))}
      {}
    &
    O(\mathit{x_a} \land \mathcal{E}_{act}(a)) \quad \text{}
  }
\]
Thus, it is not permissible for the AI system to refrain from an ethically required action.
\end{proof}
While Theorems ~\ref{theorem1} and ~\ref{theorem2} contribute to formalizing the general ethics of an AI system, it remains essential to develop rigorous formalizations for each ethical principle. 
\subsection{Formalizing Fairness} \label{formalizing fairness}
Let us begin by formalizing the concept of fairness in an AI system by considering various scenarios where it must maintain fairness and where it might compromise it. It is an important aspect of an ethical AI and can be categorized into transient fairness and stable fairness. Existing literature suggests that a fair AI system should avoid considering the sensitive attributes of individuals in its decision-making process (Definition~\ref{defn_fairness}). These attributes can potentially harm their sentiments and social standing, or even pose risks in the case of crucial applications. Such an AI system, considering sensitive attributes for making decisions, is referred to as biased and hence is not ethical~\cite{r40_biasdefn}. Hence, the following set of axioms has been developed to specify the required and complete properties for an AI system to be deemed fair. In this work, we define fairness based on the concepts outlined by Chen et al.~\cite{ftu}.
\begin{definition}[Fairness] \label{defn_fairness}
An AI system $\mathit{x}$ is fair as long as it refrains from considering sensitive attributes in the decision-making process~\cite{ftu}
\begin{enumerate}[label=(\roman*),leftmargin=2em] 
    \item \ $\mathit{x}$ exhibits stable fairness if $\Box \mathcal{F}(\mathit{x})$.
    \item \ $\mathit{x}$ exhibits transient fairness if $\mathcal{F}(\mathit{x})$ at time $t_1$ and $\neg \mathcal{F}(\mathit{x})$ at time $t_2$ where $t_1 \neq t_2$.
\end{enumerate}
\end{definition}
\footnote{An AI system is said to be \emph{unfair} if its decision-making process or its outcomes result in discriminatory treatment of individuals based on sensitive attributes, either explicitly (through direct use of such features) or implicitly (through proxies, biased data distributions, or random disparities that systematically disadvantage protected groups).
}
\begin{axiomset}{Fairness}\label{A2}
\begin{enumerate}[label=2.\arabic*,leftmargin=2em]
   \item \ \textit{Ethical AI systems exhibit stable fairness:  $ \mathcal{E}(\mathit{x}) \rightarrow \Box  (\mathcal{F}(\mathit{x}))$} \label{axiom4}
    \item \ \textit{If an AI system ever exhibits bias, it violates ethics:  $\mathcal{B}(\mathit{x}) \rightarrow \neg \mathcal{E}(\mathit{x})$}\label{axiom5}
    \item \ \textit{AI systems should not exhibit bias until ensuring fairness mechanisms are in place:  $ \neg \mathcal{B}(\mathit{x}) \, \mathcal{U} \, \mathcal{F}(\mathit{x})$} \label{axiom6}
    \item \ \textit{Fairness on the training distribution does not necessarily transfer to the deployment distribution: 
     $\neg \Box  (\mathcal{F}(\mathit{x})_{train} \rightarrow \mathcal{F}(\mathit{x})_{deploy})$} \label{axiom7}
     \item \ \textit{Lack of fairness implies the presence of bias: 
     $\neg \mathcal{F}(\mathit{x}) \rightarrow \mathcal{B}(\mathit{x})$} \label{axiomf8}
     \item \ \textit{An iterative learning system eventually mitigates bias independently in both the training and deployment phases: $ \mathcal{L}(x) \;\rightarrow\; \diamond \neg \mathcal{B}(x)_{\mathit{train}} \;\land\; \diamond \neg \mathcal{B}(x)_{\mathit{deploy}}$} \label{ax:iterative-fairness} 

\end{enumerate}
\end{axiomset}

 Axiom ~\ref{axiom4} states that, if an AI system ever commits to fairness, it should maintain this commitment throughout its usage. Let us consider that, initially, the system is trained rigorously to make decisions while being fair. However, over time, it may begin to consider sensitive attributes in its decision-making process due to skewness or disparities in real-world data. 
In such cases, the system must undergo iterative training to eliminate sensitive attributes to incorporate fairness constraints.
In some cases, even after iterative training, over time, a system may begin to consider sensitive attributes or undertake actions beyond its legal obligations. This may introduce biases by compromising its fairness. Hence Axiom~\ref{axiom5} states that in such instances, it deviates from ethical standards.
As long as an AI system maintains fairness either through iterative training or one-time training in its decision-making process, it will inherently mitigate biases, ensuring equitable treatment for all individuals. Axiom~\ref{axiom6} enforces a formal requirement: the presence of bias is disallowed prior to the establishment of fairness. Furthermore, the training distribution and deployment distribution of data are not identical in the real world. Hence, ensuring fairness in the distribution of training data does not automatically ensure fairness in the distribution of deployed systems, as real-world deployment scenarios may introduce additional biases and disparities that need to be addressed separately. This property is expressed in Axiom~\ref{axiom7}. Additionally, Axiom~\ref{axiomf8} states that a lack of fairness implies the presence of bias in the decision-making process. Finally, Axiom~\ref{ax:iterative-fairness} formalizes the assumption that iterative learning mechanisms progressively mitigate bias in both the training and deployment phases. The axiom states that if a system employs iterative learning, it will eventually reach states in which bias is eliminated during training and deployment independently. This assumption reflects the practical behavior of many modern machine learning systems, where models are repeatedly updated using new data, feedback signals, or fairness-aware optimization procedures. This axiom is motivated by Axiom~\ref{axiom7}, which states that fairness achieved on the training distribution does not necessarily transfer to the deployment environment. Because of this potential distribution shift, reducing bias during training alone is insufficient. Hence, an ethical iterative system must address both phases independently. 
\par
Based on the above foundational principles, the ethics of an AI system in terms of fairness can be formally verified using the following set of theorems. Theorem~\ref{theorem3} states that for an AI system to maintain ethical standards, it must refrain from displaying bias and consistently uphold fairness in all its operations. 

\begin{theorem} \label{theorem3}
If an AI system ever loses fairness, then it will eventually violate ethics: $\diamond \neg \mathcal{F}(\mathit{x}) \rightarrow \diamond \neg \mathcal{E}(\mathit{x})$
\end{theorem}
\begin{proof} ~

\begin{enumerate} [leftmargin=2em]

\item \ Assume $\diamond \neg \mathcal{F}(\mathit{x})$. Then, by the definition of the $\diamond$ operator, $\neg \mathcal{F}(\mathit{x})$ eventually becomes true 
\item \ From Axiom ~\ref{axiom5}, we have: $\mathcal{B}(\mathit{x}) \rightarrow \neg \mathcal{E}(\mathit{x})$
\item \ From Axiom ~\ref{axiomf8}, we have $\neg \mathcal{F}(\mathit{x})\rightarrow \mathcal{B}(\mathit{x})$ 
\item \ Combining 2 and 3 using transitivity of implication gives: $\neg \mathcal{F}(\mathit{x}) \rightarrow \neg \mathcal{E}(\mathit{x})$
\item \ By monotonicity of $\diamond$ rule, we have $\diamond \neg \mathcal{F}(\mathit{x}) \rightarrow \diamond\neg \mathcal{E}(\mathit{x})$
\end{enumerate}

\end{proof}

Given the significance of fairness in an ethical system, it is acknowledged that over time, discrepancies in data or training methods may cause the system to temporarily lose fairness, only to regain it later. In such instances, consistency cannot be guaranteed, leading to intermittent biases. However, based on Axiom~\ref{axiom4}, it is understood that once committed to acting fairly, the AI system should maintain that fairness consistently. Theorem~\ref{theorem4} captures this nuanced requirement---ethical systems must have stable fairness. This means that, if fairness is temporarily lost, systems cannot be intermittently unfair and must regain permanent fairness at some defined point. This property helps to prevent unbounded unfairness. An AI system should either consistently maintain fairness, or if unfairness exists, it should only persist until a fairness mechanism is put in place. The significance of this theorem is that it goes beyond a simple requirement of fairness and provides precise temporal constraints. Hence, it requires ethical systems to ``fix'' any temporary losses of fairness within a bounded time frame.

\begin{theorem} \label{theorem4}

An ethical AI system exhibits either stable fairness or transient fairness followed by stable fairness, but never intermittent fairness: \\
$\mathcal{E}(\mathit{x}) \rightarrow \big((\Box \mathcal{F}(\mathit{x})) \lor (\diamond \mathcal{F}(\mathit{x}) \; \land \; \Box(\neg \mathcal{F}(\mathit{x}) \mathcal{U} \mathcal{F}(\mathit{x})))\big)$
\end{theorem}
\begin{proof}~

\begin{enumerate}[leftmargin=2em]

\item \ Assume $\mathcal{E}(\mathit{x})$. \hfill (Assumption)

\item \ From Axiom~\ref{axiom4} $\mathcal{E}(\mathit{x}) \rightarrow \Box \mathcal{F}(\mathit{x})$, by modus ponens we derive:$\Box \mathcal{F}(\mathit{x})$
\item \ Since $\Box \mathcal{F}(\mathit{x})$ holds, by LTL semantics it follows that:
$\diamond \mathcal{F}(\mathit{x})$
(because what always holds trivially eventually).
\item \ From step 2, $\mathcal{F}(\mathit{x})$ holds at all time points, then the temporal Until condition is satisfied at every state:
$\Box(\neg \mathcal{F}(\mathit{x}) \mathcal{U} \mathcal{F}(\mathit{x}))$
since $\mathcal{F}(\mathit{x})$ already holds immediately.

\item \ From Steps 3 and 4, we obtain:
$\diamond \mathcal{F}(\mathit{x}) \; \land \; \Box(\neg \mathcal{F}(\mathit{x}) \mathcal{U} \mathcal{F}(\mathit{x}))$
\item \ From Step 2, we already have $\Box \mathcal{F}(\mathit{x})$.  
Hence, by disjunction introduction:
$(\Box \mathcal{F}(\mathit{x})) \lor (\diamond \mathcal{F}(\mathit{x}) \; \land \; \Box(\neg \mathcal{F}(\mathit{x}) \mathcal{U} \mathcal{F}(\mathit{x})))$

\item \ Therefore,
$\mathcal{E}(\mathit{x}) \rightarrow 
\big((\Box \mathcal{F}(\mathit{x})) \lor 
(\diamond \mathcal{F}(\mathit{x}) \; \land \; 
\Box(\neg \mathcal{F}(\mathit{x}) \mathcal{U} \mathcal{F}(\mathit{x})))\big).$

\end{enumerate}

\end{proof}
We acknowledge that fairness is a deeply contested and context-dependent concept in both ethics and AI literature, with multiple definitions often leading to trade-offs and incompatibilities. The formalization presented here captures only a subset of these ideas, tailored to settings where individual treatment based on sensitive features is of primary concern. Our goal is not to exhaustively define fairness but to demonstrate how certain ethical concerns—such as non-discrimination—can be encoded and verified within a formal framework. We leave the integration of richer, possibly conflicting notions of fairness to future work.

\subsection{Iterative Learning}
In real-world applications, many AI systems are not static but evolve. Theorem~\ref{theorem5} states that in such AI systems that learn continuously over time, fairness mechanisms have to be enforced both
during initial training and later during real-world operation. From Axiom~\ref{axiom7} it is evident that fairness in the training data does not guarantee fairness during deployment. Hence, for iterative learning systems, we must monitor for fairness issues
offline (during training) and online (during deployment). This helps in achieving stable fairness. To provide the reader with a context on iterative learning, we provide a definition (Definition~\ref{defn_iterativelearning}) grounded in domain knowledge and earlier works, as outlined by Goodfellow et al.~\cite{iterative_learning_Goodfellow-et-al-2016}.

 \begin{definition} [Iterative learning] 
 \label{defn_iterativelearning}
 
Iterative learning is a sequential learning process in which an AI system updates its internal model state over multiple iterations using newly observed data at each step. Formally, let $\theta_t$ denote the model parameters (state) at iteration $t$, and let $\omega_t$ denote the data available at iteration $t$. The learning process is defined by an update rule
\[
\theta_{t+1} = h(\theta_t, \omega_t),
\]
where $h$ is a learning or update function (e.g., gradient descent or Bayesian update). The initial state $\theta_0$ is obtained through prior knowledge or initialization, and the sequence $\{\theta_t\}$ represents the evolution of the system through iterative learning.
\end{definition}

\begin{theorem} \label{theorem5}  

For ethical iterative learning systems, the fairness constraint is eventually enforced both during training and deployment: 
     $\mathcal{E}(\mathit{x}) \land \mathcal{L}(\mathit{x}) \rightarrow  (\diamond \mathcal{F}(\mathit{x})_{train} \land \diamond \mathcal{F}(\mathit{x})_{deploy})$
\end{theorem} 

\begin{proof}~

\begin{enumerate} [leftmargin=2em]

\item \ Assume $\mathcal{E}(x) \land \mathcal{L}(x)
$ (Assumption)

\item \ From (1) by simplification: $ \mathcal{L}(x) $

\item \ From Axiom~\ref{ax:iterative-fairness}:
$\mathcal{L}(x)\rightarrow(\diamond \neg \mathcal{B}(x)_{train}
\land\diamond \neg \mathcal{B}(x)_{deploy})$

\item \ From (2) and (3) by modus ponens: $
\diamond \neg \mathcal{B}(x)_{train}\land\diamond \neg \mathcal{B}(x)_{deploy}$

\item \ From Axiom~\ref{axiomf8}: $\neg \mathcal{F}(x) \rightarrow \mathcal{B}(x) $. Taking the contrapositive: $\neg \mathcal{B}(x) \rightarrow \mathcal{F}(x)$

\item \ Therefore, $\diamond \neg \mathcal{B}(x)_{train} \rightarrow\diamond \mathcal{F}(x)_{train}$ and $
\diamond \neg \mathcal{B}(x)_{deploy}
\rightarrow
\diamond \mathcal{F}(x)_{deploy}$

\item \ From (4) and (6):$\diamond \mathcal{F}(x)_{train}
\land\diamond \mathcal{F}(x)_{deploy}$

\item \ Therefore $
\mathcal{E}(x) \land \mathcal{L}(x)\rightarrow(\diamond \mathcal{F}(x)_{train}\land\diamond \mathcal{F}(x)_{deploy}) $(1–7, Conditional Proof)

\end{enumerate}

\end{proof}

Here is an additional result based on Axiom~\ref{axiom7} examining the relationship between training and deployment for AI systems. Theorem~\ref{theorem6} states that even if bias emerges during the training phase of a system due to subjective judgments or personal opinions embedded in the data, the presence of an iterative learning mechanism enables the system to eventually mitigate such bias during deployment. In practice, achieving this bias mitigation requires the integration of fairness-aware techniques throughout the machine learning lifecycle. These typically include pre-processing approaches (e.g., data balancing or reweighting), in-processing techniques (e.g., fairness-constrained optimization), and post-processing methods (e.g., adjusting decision thresholds)~\cite{Biasmitigation}. When bias is detected during training, iterative learning, particularly through feedback loops, updated datasets, or counterfactual examples, can progressively correct it. Techniques such as adversarial debiasing~\cite{adversarial_learning}, continual fine-tuning~\cite{continual_learning}, and customized loss functions facilitate this gradual improvement, allowing the system to reduce bias and move toward fairer behavior over time.

\begin{theorem} \label{theorem6}
If an AI system exhibits bias during training but employs iterative learning, 
then, bias will eventually be eliminated during deployment:
\[
\mathcal{B}(\mathit{x})_{train} \land \mathcal{L}(\mathit{x})
\rightarrow
\diamond \neg \mathcal{B}(\mathit{x})_{deploy}
\]
\end{theorem}

\begin{proof}~

\begin{enumerate}[leftmargin=2em]

\item \ Assume  $ \mathcal{B}(\mathit{x})_{train} \land \mathcal{L}(\mathit{x})
$ (Assumption)

\item \ From (1) by simplification: $ \mathcal{L}(\mathit{x}) $

\item \ From Axiom~\ref{ax:iterative-fairness}: $ \mathcal{L}(\mathit{x}) \rightarrow
(\diamond \neg \mathcal{B}(\mathit{x})_{train} \land \diamond \neg \mathcal{B}(\mathit{x})_{deploy}) $

\item \ From (2) and (3) by modus ponens:
$ \diamond \neg \mathcal{B}(\mathit{x})_{train} \land \diamond \neg \mathcal{B}(\mathit{x})_{deploy} $

\item \ From (4)  and  (1) by simplification: $ \diamond \neg \mathcal{B}(\mathit{x})_{deploy} $

\item \ Therefore, $ \mathcal{B}(\mathit{x})_{train} \land \mathcal{L}(\mathit{x})
\rightarrow \diamond \neg \mathcal{B}(\mathit{x})_{deploy}$ (1--5, Conditional Proof)

\end{enumerate}

\end{proof}

\subsection{Formalizing Explainability } \label{formalizing explainability}
In addition to fairness, explainability is an important aspect of an ethical AI system. It is the ability of an AI system to be transparent and provide understandable explanations for its decisions. This allows individuals to understand the attributes considered in the decision-making process, enabling them to evaluate the ethical integrity of the system. There are two types of explainability in the literature: inherent and retrofitted explainability. Building upon established notions of explainability, this work defines explainability (Definition~\ref{defn_xai}) following the concept presented by Das and Rad~\cite{r39_xaidefns}. 
\begin{definition}[Explainability] \label{defn_xai}
    An AI system $\mathit{x}$ is considered explainable if it provides meta-information regarding the significance of features in the decision-making process.
    \begin{enumerate} [label=(\roman*),leftmargin=2em]
        \item \ $\mathit{x}$ exhibits retrofit explainability when it relies on an external algorithm for providing explanations.
        \item \ $\mathit{x}$ is inherently explainable if it produces explanations for its predictions, without relying on external explanation methods.
    \end{enumerate}
\end{definition}

 By providing explanations, the AI system helps individuals identify which features need modification to achieve the desired (favorable) change in prediction. This concept is commonly referred to as a counterexample or counterfactual explanation in the field of AI. Essentially, it means that while the factual outcome is the result observed, the counterfactual outcome would be the desired result. If an individual receives the counterfactual, they can determine whether sensitive features played a role in the decision. This helps in verifying the counterfactual fairness of an AI system.
\begin{definition}[Transparency] \label{defn_transparency}
   An AI system $\mathit{x}$ is transparent if it is explainable to humans~\cite{r39_xaidefns}. 
\end{definition}

\begin{axiomset} {Explainability} \label{A3}
\begin{enumerate}[label=3.\arabic*,leftmargin=2em]
    
    \item \ \textit{Transparency is a necessary condition for an AI system to be ethical: $\mathcal{E}(\mathit{x}) \rightarrow \mathcal{T}(\mathit{x})$} \label{axiom8}
    \item \ \textit{The violation of a counterfactual fairness constraint $\mathit{c}$ eventually permits representation bias: $\neg \mathcal{C}(\mathit{x},\mathit{c}) \rightarrow \diamond \mathcal{B}(\mathit{x})$} \label{axiom9}
    \item \ \textit{For any AI system \( \mathit{x} \), retrofitted explainability implies ethical compliance: 
\( \mathcal{R}(\mathit{x}) \rightarrow \mathcal{E}(\mathit{x}) \).} \label{axiom3.3}

\end{enumerate}
\end{axiomset}
Axiom~\ref{axiom8} asserts the necessity of transparency for an ethical system. Transparency enables individuals to identify the factors influencing decisions, helping them to strategically adjust these attributes and values to achieve favorable outcomes. This contributes to improving trust in the system, a crucial component of ethical operation. Furthermore, Axiom~\ref{axiom9} explains that the violation of counterfactual fairness constraints leads to representation bias. Representation bias occurs when underrepresented groups experience inaccurate outcomes due to insufficient or biased data. Enforcing counterfactual fairness constraints helps to mitigate representation bias by ensuring that the decisions made by an AI system remain consistent even when a sensitive attribute, say, $gender$, is altered. We define counterfactual fairness (Definition~\ref{defn_cffairness}) based on the concept presented by Kusner et al.~\cite{r37_counterfactualfairdefn}. By using such constraints, the system is forced to make decisions based on relevant factors that are not biased against particular groups. 
\begin{definition} [Counterfactual fairness] \label{defn_cffairness}
    An AI system $\mathit{x}$ satisfies counterfactual fairness under criterion $\mathit{c}$ iff  $\Box \,\mathcal{C}(x, c)$, where $\mathcal{C}(x, c)$ is the invariance of decisions when the sensitive attribute is altered.
\end{definition}
Theorem~\ref{theorem7} captures the requirement of explainability for ethical AI---it states that ethical AI systems must either have
inherent explainability $\mathcal{X}(\mathit{x})$ designed directly into the system, or
they must eventually be retrofitted later on to provide explainability
$\mathcal{R}(\mathit{x})$. Retrofitting explainability can be achieved through counterfactual explanations, where the system provides a counterexample for changing the outcome to the desired one.

\begin{theorem} \label{theorem7}
Ethical AI systems should eventually exhibit either inherent 
explainability or retrofitted explainability:
$\mathcal{E}(\mathit{x}) \rightarrow \diamond 
(\mathcal{X}(\mathit{x}) \lor \mathcal{R}(\mathit{x}))$
\end{theorem}

\begin{proof}~

\begin{enumerate}[leftmargin=2em]

    \item \ Assume $\mathcal{E}(\mathit{x})$. 
    (Conditional proof assumption)

    \item \ From Axiom~\ref{axiom8}, we have: 
    $\mathcal{E}(\mathit{x}) \rightarrow \mathcal{T}(\mathit{x})$

    \item \ By modus ponens on Steps 1 and 2: 
    $\mathcal{T}(\mathit{x})$

    \item \ From Definition~\ref{defn_transparency}, 
    the contrapositive gives: 
    $\lnot(\mathcal{X}(\mathit{x}) \lor \mathcal{R}(\mathit{x})) 
    \rightarrow \lnot\mathcal{T}(\mathit{x})$

    \item \ By contraposition on Step 4: 
    $\mathcal{T}(\mathit{x}) \rightarrow 
    (\mathcal{X}(\mathit{x}) \lor \mathcal{R}(\mathit{x}))$

    \item \ By modus ponens on Steps 3 and 5: 
    $\mathcal{X}(\mathit{x}) \lor \mathcal{R}(\mathit{x})$

    \item \ By LTL tautology 
    $\varphi \rightarrow \diamond\varphi$ 
    (if $\varphi$ holds now, it holds eventually), 
    applied to Step 6:
    $(\mathcal{X}(\mathit{x}) \lor \mathcal{R}(\mathit{x})) 
    \rightarrow 
    \diamond(\mathcal{X}(\mathit{x}) \lor \mathcal{R}(\mathit{x}))$

    \item \ By modus ponens on Steps 6 and 7: 
    $\diamond(\mathcal{X}(\mathit{x}) \lor \mathcal{R}(\mathit{x}))$

    \item \ By conditional proof, discharging 
    assumption in Step 1:
    $\mathcal{E}(\mathit{x}) \rightarrow 
    \diamond(\mathcal{X}(\mathit{x}) \lor \mathcal{R}(\mathit{x}))$
\end{enumerate}
\end{proof}

\begin{theorem} \label{theorem8}
An ethical AI system that does not satisfy the  counterfactual fairness eventually violates ethics: 
     $\mathcal{E}(\mathit{x}) \land \diamond \neg \mathcal{C}(\mathit{x},\mathit{c}) \rightarrow \diamond \neg \mathcal{E}(\mathit{x})$.
\end{theorem}
\begin{proof} ~
 
\begin{enumerate}[leftmargin=2em]

    \item \ Assume $\mathcal{E}(\mathit{x}) \land \diamond \neg \mathcal{C}(\mathit{x},\mathit{c})$
    \item \ By temporal semantics, there exist future time $t$ where $\neg \mathcal{C}(\mathit{x},\mathit{c})$ holds
    \item \ From Axiom~\ref{axiom9}, $\neg \mathcal{C}(\mathit{x},\mathit{c}) \rightarrow \diamond \mathcal{B}(\mathit{x})$   
    \item \ From steps 2 and 3, $\diamond \mathcal{B}(\mathit{x})$
    \item \ From Axiom~\ref{axiom5}, we have: $\mathcal{B}(\mathit{x}) \rightarrow \neg \mathcal{E}(\mathit{x})$
    \item \ By applying Rule of Necessitation, $\Box(\mathcal{B}(\mathit{x}) \rightarrow \neg \mathcal{E}(\mathit{x}))$
    \item \ By applying monotonicity of $\diamond$ to step 6, $\diamond \mathcal{B}(\mathit{x}) \rightarrow \diamond \neg  \mathcal{E}(\mathit{x})$
    \item \ From steps 4 and 7 by Modus Ponens,  $\diamond \neg \mathcal{E}(\mathit{x})$
    \item \ Therefore, $\mathcal{E}(\mathit{x}) \land \diamond \neg \mathcal{C}(\mathit{x},\mathit{c}) \rightarrow \diamond \neg \mathcal{E}(\mathit{x})$

\end{enumerate}

\end{proof} 

Theorem~\ref{theorem8} formally relates
counterfactual fairness to ethical integrity.  It follows from Axiom~\ref{axiom9} and establishes that if an ethical AI system eventually \emph{violates} a counterfactual fairness constraint~$c$, then the system will eventually lose its ethical status.  The key intuition is the contrapositive reading: for an AI system to \emph{remain} ethical over time, it must consistently satisfy counterfactual fairness constraints.  This provides a formal incentive for enforcing such constraints, since any lapse eventually undermines
the system's ethical standing.
\section{Verification} \label{implementation}
Building on the formal framework introduced in Section~\ref{deontic logic for reasoning}, this section demonstrates its practical applicability by verifying ethical aspects of real-world AI systems---Loan prediction~\cite{r28_loanprediction} and COMPAS~\cite{r_27evaluatingCOMPAS,r_26propublicaCOMPAS}. 
For each system, various properties have been formulated to verify and ensure ethics. These properties are logical formulas of system-level specifications that address various aspects of fairness and explainability. They are designed specifically for each AI system and can differ depending on the specific task or the nature of the system.  
The verification process yields either a satisfiability or unsatisfiability response, indicating whether the system fulfills the property. If the property does not hold across all system executions, a counterexample is generated. The verification is implemented using an open-source theorem prover called Z3. It is a Satisfiability Modulo Theory (SMT) solver that is used to check the satisfiability of the logical formulas~\cite{r2_z3}. While several formal verification techniques exist—such as model checking~\cite{model_checking} and process algebra~\cite{process_algebra}—we employ theorem proving using Z3 in this work. Z3 offers the symbolic expressiveness required to encode deontic-temporal logic, which is essential for modeling ethical constraints like obligations, fairness, and permissions over time. Its support for satisfiability modulo theories enables efficient reasoning over complex properties, making it a suitable choice for our framework’s needs. 

\begin{figure*}[t]
    \centering
    \includegraphics[width=0.85\linewidth]{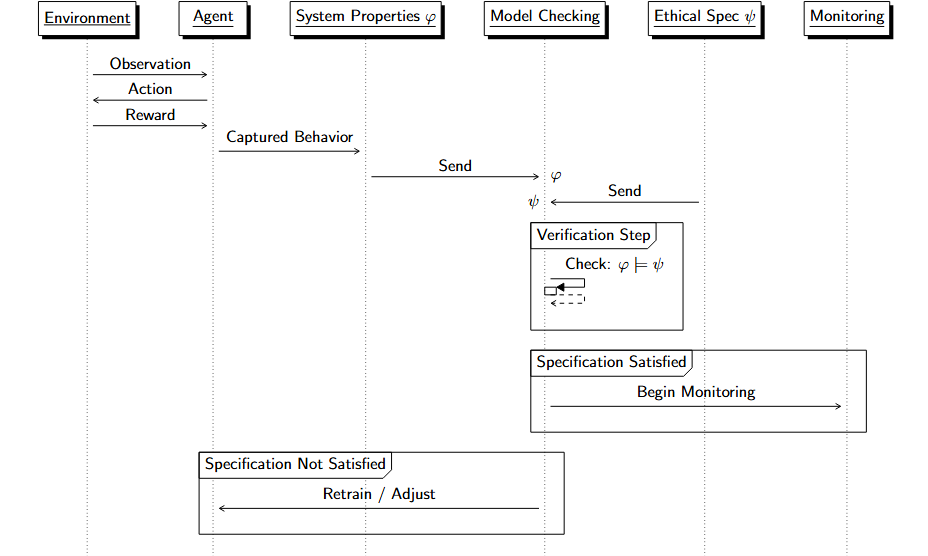}
    \caption{Sequence Diagram of Ethical Verification using Z3 Theorem Prover}
    \label{fig:z3-seq-diagram}
\end{figure*}

We emphasize that the verification performed in this work is
\emph{bounded SMT-based satisfiability checking}, not full temporal model checking in the classical sense (i.e., exhaustive state-space exploration of a Kripke structure).  The label ``Model Checking'' in the sequence diagram (Figure~\ref{fig:z3-seq-diagram}) refers to the general activity of checking whether observed system behavior satisfies a formal specification. The underlying engine is
the Z3 SMT solver.  In particular, our framework reasons over
\emph{recorded decision outcomes} in a dataset rather than over the internal architecture or learned weights of the underlying machine learning model. 
 \par
 The system incorporates a set of formal properties, denoted by the ethical specification $\psi$, which define the desired ethical behavior of the model. The actual behavior of the model, as captured during its interaction with the environment, is represented by $\varphi$. Verification is performed by checking the logical entailment $\varphi \models \psi$, determining whether the observed behavior adheres to the ethical specification. If the condition is satisfied, the system transitions to a \textit{Monitoring} phase to ensure ongoing compliance during deployment. If the specification is violated, the system initiates a \textit{Retraining} process, wherein corrective feedback is used to adjust the agent’s policy or behavior. This iterative loop facilitates continuous ethical alignment of the AI system over time.

\par
The predicates in Table~\ref{table-predicatesimpln} are used to assist in the formalization of both systems. These predicates return either true or false. Here, variables $i$ and $j$ denote individuals within these systems, each characterized by a vector representation, i.e., $i=(i_1, i_2, ..., i_m)$ where $m$ signifies the number of attributes representing an individual and $i_k$ where $k=1...m$ represents the value of the respective attribute.

\par

Algorithm~\ref{algm1} describes the verification procedure for various properties of the COMPAS system, while Algorithm~\ref{algm2} pertains to the loan prediction system. These algorithms provide clear insights into the implementation of the framework in real-world AI systems. In both algorithms, a dataset containing information about individuals (attributes 
 and values) and the properties formulated using deontic logic serves as input. These properties are the ethical properties that an ethical AI system should satisfy. The output, determined by the Z3 SMT solver, is a result indicating whether the specification/property is satisfiable (SAT) or unsatisfiable (UNSAT).

\begin{table}

\caption{Predicates for Ethics Evaluation in Loan Prediction and COMPAS AI}
\label{table-predicatesimpln}
\centering
\begin{tabular}
{ >{\raggedright\arraybackslash}p{2.3cm}
  >{\raggedright\arraybackslash}p{1cm}
  >{\raggedright\arraybackslash}p{4.5cm} }

\hline
AI System & Predicate & Explanation \\
\hline

\multirow{7}{*}{Loan Prediction}
& $\alpha(x,i)$ & System $x$ determines that the credit score of individual $i$ is above the required threshold \\

& $\gamma(x,i)$ & System $x$ determines that the income of individual $i$ is above the required threshold \\

& $\theta(x,i)$ & Individual $i$ applies for a loan evaluated by system $x$ \\

& $\delta(x,i)$ & System $x$ approves the loan application of individual $i$ \\

& $S(x,i,j)$ & In system $x$, individuals $i$ and $j$ have similar financial attributes \\

& $\beta(x,i)$ & System $x$ permits individual $i$ to appeals against the decision\\

& $\eta(x,i)$ & System $x$ considers sensitive attributes of individual $i$ \\

\hline

\multirow{5}{*}{COMPAS}
& $\sigma(x,i)$ & System $x$ labels individual $i$ as a recidivist \\

& $\beta(x,i)$ & System $x$ permits individual $i$ to appeals against the decision \\

& $\rho(x,i)$ & System $x$ records that individual $i$ has prior offenses \\

& $\lambda(x,i)$ & System $x$ performs recidivism assessment for individual $i$ \\

& $\eta(x,i)$ & System $x$ considers sensitive attributes of individual $i$ \\

\hline
\end{tabular}
\end{table}

Table~\ref{table-predicatesdatasetmapping} specifies how abstract logical predicates are instantiated using observable dataset attributes and decision outcomes. These tables define the formal abstraction used for ethical verification and do not represent or attempt to reconstruct the internal implementation of the underlying machine learning models. All predicates are grounded at the dataset and decision level, enabling reasoning over observable behavior.
\par
For the loan prediction system, the decision outcome is binary, where a value of $1$ denotes loan approval and a value of $0$ denotes loan rejection. The predicate $\delta(x,i)$ represents this outcome and is derived directly from the dataset’s approval label. Sensitive attributes, such as gender, are captured by the predicate $\eta(x,i)$ and are used exclusively for evaluating ethical properties, including disparate treatment and disparate impact. Similarly, for the COMPAS dataset, the decision outcome is binary and obtained by thresholding the risk score: a \texttt{decile\_score} greater than or equal to 5 is interpreted as a prediction of recidivism, while scores below 5 indicate non-recidivism. This threshold groups the COMPAS medium- and high-risk categories (scores 5–10) as positive predictions and treats scores 1–4 (low risk) as negative, following the standard COMPAS risk categorization and prior work~\cite{compas_score_threshold,compas_score_threshold1,compas_score_threshold2}. 

The predicate $\sigma(x,i)$ corresponds to this decision outcome and is instantiated using the COMPAS risk assessment labels available in the dataset, while the sensitive attribute predicate $\eta(x,i)$ captures race, with particular focus on individuals identified as African-American. In this study, both models are evaluated using example datasets obtained from the Kaggle platform and the UCI Machine Learning Repository to demonstrate the effectiveness of the proposed ethical verification framework.

\begin{table}

\caption{Grounding of Predicate Symbols to Dataset Features}
\label{table-predicatesdatasetmapping}
\centering
\begin{tabular}
{ >{\raggedright\arraybackslash}p{2.3cm}
  >{\raggedright\arraybackslash}p{1cm}
  >{\raggedright\arraybackslash}p{4.5cm} }

\hline
AI system & Predicate & Dataset Feature Mapping \\
\hline

\multirow{7}{*}{Loan Prediction}

& $\alpha(x,i)$ & \texttt{credit\_score($i$) $\geq \tau_c$} \\

& $\gamma(x,i)$ & \texttt{income($i$) $\geq \tau_{inc}$} \\

& $\theta(x,i)$ & \texttt{True (Implicit by presence in dataset)} \\

& $\delta(x,i)$ & \texttt{approved($i$) == 1} \\

& $S(x,i,j)$ & $\big(|\text{credit\_score}(i) - \text{credit\_score}(j)| \leq \epsilon_c \;\land\;
|\text{income}(i) - \text{income}(j)| \leq \epsilon_{inc}\big)$ \\

& $\eta(x,i)$ & \texttt{gender($i$) == 'Female'} \\

\hline

\multirow{5}{*}{COMPAS}

& $\sigma(x,i)$ & \texttt{decile\_score($i$) $\geq 5$} \\

& $\rho(x,i)$ & \texttt{priors\_count($i$) $> 0$} \\

& $\lambda(x,i)$ & \texttt{is\_recid($i$) == 1} \\

& $\eta(x,i)$ & \texttt{race($i$) == 'African-American'} \\

\hline
\end{tabular}
\end{table}
\subsection{COMPAS} \label{subsection_compas}
It is an AI system used in the criminal justice system to assess the likelihood of a defendant re-offending based on various factors. The input can be some features or attributes, including \textit{name, age, gender, race, address, previous criminal activities, number of years of punishment}, etc. Based on these features, the system makes a decision. This kind of decision is very critical because a wrong decision in this case will damage the social status of an individual and his/her emotional state.  Making decisions based on \textit{previous criminal activities, number of years of punishment}, and other relevant features is considered ethical. Conversely, it is unethical to use \textit{race, gender}, or \textit{age} as factors in decision-making processes. To formulate and verify the ethics of this system logically, we represent the properties as follows: \hfill
\par

\begin{proposition}{Ethics Properties for COMPAS}

\begin{enumerate}[label=(\alph*),leftmargin=2em]

\item Always it is permissible for system $x$ to assess recidivism when prior offenses exist:
\[
\forall i \; \Box P(\rho(x,i) \rightarrow \lambda(x,i))
\]
\label{compasaxiom1}

\item It is obligatory for system $x$ to conduct recidivism assessments without using sensitive attributes:
\[
\forall i \; O(\neg \eta(x,i) \rightarrow \lambda(x,i))
\]
\label{compasaxiom2}

\item It is forbidden for system $x$ to label individuals as recidivists based on sensitive attributes:
\[
\forall i \; \neg P(\eta(x,i) \rightarrow \sigma(x,i))
\]
\label{compasaxiom3}

\item It is not permitted for system $x$ to label an individual as a recidivist without prior offenses:
\[
\forall i \; \neg P(\neg \rho(x,i) \rightarrow \sigma(x,i))
\]
\label{compasaxiom4}

\item If an individual is labeled as a recidivist, it is always permitted that they can eventually appeal:
\[
\forall i \; \Box(\sigma(x,i) \rightarrow \Diamond P(\beta(x,i)))
\]
\label{compasaxiom5}

\end{enumerate}
\end{proposition}
In this scenario, the property~\ref{compasaxiom1} explains the fact that the AI system used to automate judicial recidivism is legally bound to assess the risk of re-offending crimes of all individuals with prior offenses. This ethical action aligns with Theorems (\ref{theorem1}) and (\ref{theorem2}). Property \ref{compasaxiom2} asserts that the system must refrain from evaluating the risk of the individuals by considering sensitive attributes such as $race, gender$, or $age$. This principle safeguards the fairness of the system, as defined by Theorem \ref{theorem3}. The property~\ref{compasaxiom3} explains that the outcome of the system should be consistent irrespective of alterations in the values of sensitive attributes. According to Theorem ~\ref{theorem8}, this property needs to be satisfied by an ethical system. The property~\ref{compasaxiom4} encodes that an individual should not be mislabelled as a recidivist without any prior offenses (encodes the need for fairness in the decision---Theorem~\ref{theorem3}) and finally property~\ref{compasaxiom5} explains the fact that the individual has all the legal right to question the AI system if the decision given is not acceptable for them (encodes the need for explainability---Theorem~\ref{theorem7}).
For a system to be deemed ethical, it must fulfill all properties. However, COMPAS fails to satisfy properties~\ref{compasaxiom2},~\ref{compasaxiom3}, and~\ref{compasaxiom4} thereby failing to be considered ethical. Figure~\ref{fig:correctness}, demonstrates simple yet non-trivial verification of the correctness of property~\ref{compasaxiom2}. The proof relies on contradiction, considering the negation of the property to be proved correct. It indicates that this negation violates certain axioms or theorems of ethical systems discussed in Section~\ref{deontic logic for reasoning}. Thus, the property must always be satisfied by an ethical system.
\par
Algorithm~\ref{algm1} begins with initializing the solver in line 2. Lines 4--6 declare the abstract domains required for reasoning, namely the sorts System, Person, and Time, which allow the formulation of properties over individuals and temporal instances. Line 10 introduces a binary predicate defining a partial ordering over time, enabling temporal reasoning.
Subsequently, lines 12--16 declare predicates, each defined as a Boolean-valued function over a person and a time point, to represent the system state during the process. The predicates $\rho(), \sigma(), \lambda(), \eta()$, and $\beta()$, each evaluates to either true or false. Here, $x$ represents the data distribution of the AI system and $i$ represents the specific individual for whom the properties are being tested. Lines 18 and 19 define the deontic operators $\textit{Perm}$ (permission) and $\textit{Obl}$ (obligation), which take a Boolean condition and a time instant as arguments and evaluate to a Boolean value, thereby enabling the expression of deontic constraints in a temporal context. Lines 30--32 illustrate the first property of the COMPAS system, formulated in Z3 Python code, and line 33 employs the solver instance to ascertain its satisfiability. In our example, this yields a 'satisfied' result. Similarly, lines 34--51 assess the remaining four properties, with 'satisfied' results obtained for property 5, while properties 2, 3, and 4 return 'unsatisfied'. This indicates that the system fails to adhere to some of the defined ethical properties and, therefore, is deemed unethical.

\begin{algorithm}

\caption{Z3 algorithm to check the satisfiability of deontic--temporal properties for the COMPAS AI system}
\label{algm1}
\begin{algorithmic}[1]

\Require $x$: dataset of individuals evaluated by the AI system,
$\varphi_i$: deontic--temporal property from $\varphi=[\varphi_1,\varphi_2,\varphi_3,\varphi_4,\varphi_5]$

\Ensure \textbf{Sat} if $\varphi_i$ is satisfiable, \textbf{Unsat} otherwise

\State $S \gets \text{Solver}()$ \Comment{Initialize solver}
 
\State \textbf{declare-sort} \textit{System} \Comment{Declare domains}
\State \textbf{declare-sort} \textit{Person}
\State \textbf{declare-sort} \textit{Time}

\State \textbf{declare-const} $x$ \textit{System} \Comment{Declare COMPAS system}

\State \textbf{declare-fun} $\geq(\textit{Time},\textit{Time})$ returns Bool \Comment{Declare time ordering}

\State \Comment{Declare state predicates (system, individual, time)}
\State \textbf{declare-fun} $\rho(\textit{System},\textit{Person},\textit{Time})$ returns Bool
\State \textbf{declare-fun} $\sigma(\textit{System},\textit{Person},\textit{Time})$ returns Bool
\State \textbf{declare-fun} $\lambda(\textit{System},\textit{Person},\textit{Time})$ returns Bool
\State \textbf{declare-fun} $\eta(\textit{System},\textit{Person},\textit{Time})$ returns Bool
\State \textbf{declare-fun} $\beta(\textit{System},\textit{Person},\textit{Time})$ returns Bool

\State \Comment{Declare deontic predicates}
\State \textbf{declare-fun} $Perm(\textit{Bool},\textit{Time})$ returns Bool
\State \textbf{declare-fun} $Obl(\textit{Bool},\textit{Time})$ returns Bool

\ForAll{individual $i$ in $x$} \Comment{Encode dataset $x$}
    \ForAll{time $t$}
        \State assert $\rho(x,i,t) \leftrightarrow (\text{priors\_count}(i) > 0)$
        \State assert $\sigma(x,i,t) \leftrightarrow (\text{decile\_score}(i) \geq 5)$
        \State assert $\lambda(x,i,t) \leftrightarrow (\text{is\_recid}(i) == 1)$
        \State assert $\eta(x,i,t) \leftrightarrow (\text{race}(i) == \text{'African-American'})$
        \State assert $\beta(x,i,t) \leftrightarrow (\text{score\_text}(i) == \text{'High'})$
    \EndFor
\EndFor

\ForAll{individual $i$ and time $t$} \Comment{Property $\varphi_1$}
    \State assert $Perm(Implies(\rho(x,i,t) , \lambda(x,i,t)),t)$ 
\EndFor
\State $S.\text{check}()$

\ForAll{individual $i$ and time $t$} \Comment{Property $\varphi_2$}
    \State assert $Obl(Implies(\neg\eta(x,i,t) , \lambda(x,i,t)),t)$ 
\EndFor
\State $S.\text{check}()$

\ForAll{individual $i$ and time $t$} \Comment{Property $\varphi_3$}
    \State assert $\neg Perm(Implies(\eta(x,i,t), \sigma(x,i,t)), t)$ 
\EndFor
\State $S.\text{check}()$

\ForAll{individual $i$ and time $t$} \Comment{Property $\varphi_4$}
    \State  assert $\neg Perm(Implies(\neg \rho(x,i,t), \sigma(x,i,t)), t)$ 
\EndFor
\State $S.\text{check}()$

\ForAll{individual $i$ and time $t$} \Comment{Property $\varphi_5$}
    \State \textbf{if} $\sigma(x,i,t)$ \textbf{then}
        \State \quad \textbf{exists} $t'$ such that $(t' \geq t)$ and $Perm(\beta(x,i,t'),t')$ 
    \State \textbf{end if}
\EndFor
\State $S.\text{check}()$

\end{algorithmic}
\end{algorithm}

\begin{figure}  
    \centering
         
$ \varphi_2 = (Forall \ i, 
            Implies(\neg \eta(x,i),\lambda(x,i)))$   \Comment{Z3 encoding}
\[
  \infer[\text{mp}]{\lnot \mathcal{E}(\mathit{x})}
  {
    \infer[\text{}]{\lnot \mathit{\varphi_2}}
    {
      \text{}
    }
    &
    \infer[\text{trans}]{\lnot \mathit{\varphi_2}=\lnot \mathcal{E}(\mathit{x})}
    {
      \infer[\text{mono}]{\lnot \mathit{\varphi_2} = \lnot \mathcal{F}(\mathit{x})}
{
      \text{}
    }
    &
    \infer[\text{~\ref{theorem3}}]{\lnot \mathcal{F}(\mathit{x})=\lnot \mathcal{E}(\mathit{x})}
      {
        \infer[\text{rewrite}]{\mathit{\varphi_2}=\mathcal{F}(\mathit{x}) }{
        }
        &
        \text{}
      }
    }
  }
  \]            

\caption{Correctness of Property~\ref{compasaxiom2}---COMPAS AI system}
\label{fig:correctness}
\end{figure}
  
\subsection{Loan Prediction System} \label{subsection_loan}
In automating the loan decision process, an AI system analyzes the trained data to decide on acceptance or rejection. This data includes various attributes such as \textit{income, credit score, age, occupation, education, name, address,} and \textit{gender}, although not all are relevant for decisions. Notably, features like \textit{gender} and \textit{name} are not necessary for decision-making. However, there is a risk that the system might erroneously consider these irrelevant details, which is deemed unfair or unethical. Conversely, ethical behavior is exhibited when the system considers features such as \textit{income, credit score, occupation,} and \textit{age}. We encoded this scenario using deontic logical formulas, helping us to verify the ethics of the system.

\begin{proposition}{Ethics Properties of Loan Prediction}

\begin{enumerate}[label=(\alph*),leftmargin=2em]

\item \ An AI system $x$ is obliged to provide a decision for every individual who applies for a loan:
\[
\forall i \; \Box \big( \theta(x,i) \rightarrow O(\delta(x,i) \lor \neg \delta(x,i)) \big)
\]
\label{lpaxiom1}

\item \ It is necessary that an individual meeting financial criteria is approved by the system:
\[
\forall i \; \Box \big( (\alpha(x,i) \lor \gamma(x,i)) \rightarrow \delta(x,i) \big)
\]
\label{lpaxiom2}

\item \ It is obligatory for the system to ensure fairness by assigning the same outcome to similar individuals:
\[
\forall i,j \; O \big(  S(x,i,j) \rightarrow 
((\delta(x,i) \land \delta(x,j)) \lor (\neg \delta(x,i) \land \neg \delta(x,j))) \big) 
\]
\label{lpaxiom3}

\item \ It is forbidden for the system to base decisions on sensitive attributes:
\[
\forall i \; \neg P\big( \eta(x,i) \rightarrow \delta(x,i) \big)
\]
\label{lpaxiom4}

\item \ It is always required that the system provide an appeal mechanism for rejected individuals:
\[
\forall i \; \Box \big( \neg \delta(x,i) \rightarrow \Diamond P(\beta(x,i)) \big)
\]
\label{lpaxiom5}

\end{enumerate}

\end{proposition}
 The property~\ref{lpaxiom1} encodes that the AI system is obligated to provide a decision regarding an application submitted by an individual. This defines the action an AI system should perform. By Theorems~\ref{theorem1} and ~\ref{theorem2}, this property must be upheld, and it is indeed upheld in this system. The property~\ref{lpaxiom2} states that a person should have a good credit score or income as given by the regulation to get acceptance. The threshold here specifies the lower bound set by the regulatory body. This represents the fact that the system should consider general attributes in the decision-making process. The system satisfies this property, thereby upholding Theorem~\ref{theorem3}.
 The property~\ref{lpaxiom3} emphasizes the importance of upholding equality or fairness, where two similar individuals should receive similar decisions. This property encodes that no discrimination should be there, and the decision made by the system should be consistently based on only ethical actions, and hence, by Theorem~\ref{theorem3}, it cannot be violated. However, our empirical analysis of the loan dataset indicates that the learned decision outcomes do not uphold this property, revealing a gender-based disparity in decisions within the analyzed dataset.
 The property~\ref{lpaxiom4} explains that the system must maintain consistency in its decisions over time, without altering them based on sensitive features. However, the system fails to adhere to this property, thereby violating ethics according to Theorem \ref{theorem8}.
 The property~\ref{lpaxiom5} states the need for explainability, ensuring that individuals have the legal right to question decisions and receive valid explanations. Failure to provide such explanations violates Theorem \ref{theorem7}.  While the system meets this property, it still fails to satisfy all five properties, thereby failing to be considered ethical. 
\begin{algorithm}

\caption{Z3 algorithm to check the satisfiability of deontic--temporal properties for the Loan Prediction AI system}
\label{algm2}
\begin{algorithmic}[1]

\Require $x$: Data distribution of the AI system, $\varphi_i$: quantified deontic--temporal property from the list $\varphi=[\varphi_1,\varphi_2,\varphi_3,\varphi_4,\varphi_5]$ \Ensure \textbf{Sat} if $\varphi_i$ is satisfiable for $x$, \textbf{Unsat} otherwise

\State $S \gets \text{Solver}()$ \Comment{Initialize solver}

\State \textbf{declare-sort} \textit{System}, \textit{Person}, \textit{Time} \Comment{Domains}
\State \textbf{declare-const} $x$ : \textit{System} \Comment{Loan system}

\State \textbf{declare-fun} $\geq(\textit{Time},\textit{Time})$ : Bool \Comment{Time ordering}

\State \textbf{declare-fun} $\theta, \alpha, \gamma, \delta, \eta(\textit{System},\textit{Person},\textit{Time})$ : Bool \Comment{State predicates}
\State \textbf{declare-fun} $\beta(\textit{System},\textit{Person},\textit{Time})$ : Bool \Comment{Appeal by system}
\State \textbf{declare-fun} $S(\textit{System},\textit{Person},\textit{Person},\textit{Time})$ : Bool \Comment{Similarity}

\State \textbf{declare-fun} $Perm, Obl(\textit{Bool},\textit{Time})$ : Bool \Comment{Deontic operators}

\ForAll{individual $i \in x$}
    \ForAll{time $t$}
        \State assert $\theta(x,i,t) \leftrightarrow \text{applied}(i)$
        \State assert $\alpha(x,i,t) \leftrightarrow (\text{credit\_score}(i) \geq \tau_c)$
        \State assert $\gamma(x,i,t) \leftrightarrow (\text{income}(i) \geq \tau_{inc})$
        \State assert $\delta(x,i,t) \leftrightarrow (\text{approved}(i) = 1)$
        \State assert $\eta(x,i,t) \leftrightarrow (\text{gender}(i) = \text{'Female'})$
        \State assert $\beta(x,i,t) \leftrightarrow (\text{approved}(i) = 0)$
    \EndFor

    \ForAll{individual $j \in x$ and time $t$}
        \State assert $S(x,i,j,t) \leftrightarrow \big(|\text{credit\_score}(i) - \text{credit\_score}(j)| \leq \epsilon_c \;\land\;
|\text{income}(i) - \text{income}(j)| \leq \epsilon_{inc}\big)$
    \EndFor
\EndFor

\ForAll{individual $i$, time $t$} \Comment{$\varphi_1$}
    \State assert $Implies(\theta(x,i,t), \; Obl(\delta(x,i,t) \lor \neg\delta(x,i,t), t))$
\EndFor
\State $S.\text{check}()$

\ForAll{individual $i$, time $t$} \Comment{$\varphi_2$}
    \State assert $Implies((\alpha(x,i,t) \lor \gamma(x,i,t)), \delta(x,i,t),t)$
\EndFor
\State $S.\text{check}()$

\ForAll{individuals $i,j$, time $t$} \Comment{$\varphi_3$}
   \State assert $Implies(Sim(x,i,j,t),$
    \State \hspace{1cm} $Obl((\delta(x,i,t) \land \delta(x,j,t)) \lor (\neg\delta(x,i,t) \land \neg\delta(x,j,t)), t))$

\EndFor
\State $S.\text{check}()$

\ForAll{individual $i$, time $t$} \Comment{$\varphi_4$}
    \State assert $\neg Perm(Implies(\eta(x,i,t), \delta(x,i,t)), t)$
\EndFor
\State $S.\text{check}()$

\ForAll{individual $i$, time $t$} \Comment{$\varphi_5$}
    \State \textbf{if} $\neg\delta(x,i,t)$ \textbf{then}
        \State \quad \textbf{exists} $t' \geq t$ such that $Perm(\beta(x,i,t'), t')$
\EndFor
\State $S.\text{check}()$

\end{algorithmic}
\end{algorithm}

 Algorithm~\ref{algm2} outlines the verification process for assessing the ethical properties of a loan prediction system. Five properties are formulated to evaluate the ethics of the system. Similar to the previous COMPAS algorithm, variables $x$, $i$, and $j$ represent the data distribution of the AI system and the individuals, respectively, and various functions are declared to represent predicates necessary for formulating each property (lines 2--21). The details of the predicates are given in Table~\ref{table-predicatesimpln}. Lines 22--24 encode the first property to be verified, and line 25 employs the Z3 solver instance to verify it, returning a 'satisfied' result for our dataset. Similarly, lines 26--43 handle the verification of the remaining four properties. Our solver returns 'satisfied' for properties 1, 2, and 5, while properties 3 and 4 are  'unsatisfied'.
\par
The results obtained from the simulation of these two scenarios in the Z3 theorem prover for the above-defined properties are given in Table~\ref{table-result}. The results indicate the satisfiability or unsatisfiability of the properties for the two example AI systems. From the result, it is clear that the example AI systems fail to satisfy certain properties and hence are not ethical systems. To be precise, they are using sensitive features in their decision-making process quite often and hence cannot be fair to the human population using the system. Furthermore, this result confirms the utility of our formalization method in verifying the ethical properties of any AI system. By formulating specific properties for testing, we can ascertain whether these properties are satisfiable or unsatisfiable. 
\begin{table}
  \caption{Z3 verification result obtained on two examples.}
  \label{table-result}
  \centering
  \begin{tabular}
  { >{\raggedright\arraybackslash}p{1.5cm}    >{\raggedright\arraybackslash}p{3cm}  >{\raggedright\arraybackslash}p{3cm}}  
    \toprule
     
   Property & Loan prediction & COMPAS \\
   \midrule
   \textit{a}  & Satisfied & Satisfied\\
  \textit{b}  & Satisfied & Not Satisfied\\
   \textit{c}  & Not Satisfied & Not Satisfied\\
    \textit{d}  & Not Satisfied & Not Satisfied \\
   \textit{e}  & Satisfied & Satisfied\\
    \bottomrule
  \end{tabular}
\end{table}

\begin{figure}
    \centering
    \includegraphics[width=8cm,height=4cm]{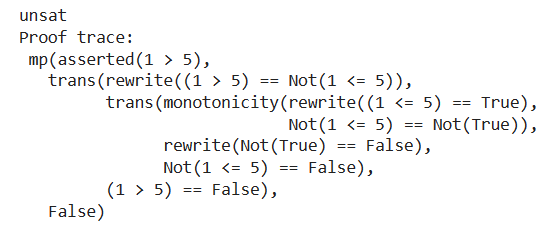}
    \caption{\textbf{A proof trace in Z3.} The property~\ref{compasaxiom2} is rewritten as the Not(negation). Using monotonicity and transitivity, it is proved that the Not(negation) is false, and hence the original condition asserted is also false. This yields an 'unsatisfied' result in the solver.  }
    \label{fig:prooftrace}
\end{figure}

Figure~\ref{fig:prooftrace} is an example Z3 proof trace demonstrating the validation of Property~\ref{compasaxiom2} of the COMPAS AI system while testing with the dataset. This example demonstrates how a property formulated in the deontic temporal formalism is verified using Z3, highlighting the effectiveness of the formalization. The property states that an AI system should consider non-sensitive attributes for making decisions.  

When iterating the property in Z3 for a randomly selected individual $\mathit{i}$ with an outcome of '1' (indicating a prediction of recommitting a crime), and considering the non-sensitive attribute $Decile-score$ from the dataset, which ranges from '0' to '10' (with '10' representing the highest risk), we encounter a discrepancy.  $Decile-score$ indicates whether the individual has a risk of re-committing the crime. According to the property, if the $Decile-score$ is greater than or equal to '5' (here '5' is the threshold considered), the outcome can be '1'. However, for this individual, the $Decile-score$ from the dataset is '1', yet the outcome remains '1'. This violation of the property leads to an 'unsatisfied' result in the verification process, as it fails to hold for the entire dataset.
\subsection{Computational Complexity and Scalability}

Although a comprehensive quantitative evaluation of computational complexity and scalability lies beyond the scope of this study (see Subsection~\ref{limitations}), a brief qualitative discussion is appropriate. The practical deployment of the proposed framework requires attention to computational constraints. The complexity of verifying deontic–temporal specifications depends on the underlying logical foundations. For specifications formulated in Linear Temporal Logic (LTL), the satisfiability problem is PSPACE-complete in the size of the formula, while model checking remains polynomial in the size of the transition system~\cite{ltl_model_checking_pspace_complete}. When deontic operators are integrated with temporal modalities, in combined temporal–normative logics, the overall complexity may increase depending on the expressiveness of the interaction between the modalities; in general formulations, decision problems have been shown to range from PSPACE up to EXPTIME-hard~\cite{agotnes2007temporal}.

From a scalability perspective, verification remains feasible for moderately large systems when the decision problem is PSPACE-bounded. However, in more expressive temporal–deontic formulations where the complexity rises to EXPTIME-hard, the worst-case verification cost increases substantially, potentially limiting scalability for highly complex normative specifications. Strategies to mitigate these limitations are discussed briefly in future work.

\subsection{Scope and Limitations}\label{limitations}
All results reported in this work are based on example datasets used for evaluation purposes. The identified biases in the COMPAS and loan prediction case studies are not claims about the correctness or intent of deployed systems, but serve to illustrate how the proposed framework can detect and reason about ethical violations when they are present in observable decision data.

More precisely, the verification operates at the level of
observable inputs, outputs, and metadata recorded in the evaluation datasets.  We do not verify the properties of the internal model weights or of the training procedure itself; instead, we assess whether the \emph{externally observable decisions} produced by the system satisfy or violate the specified ethical properties. This distinction is important: a property such as ``the system does not consider sensitive attributes'' is checked against the recorded decision outcomes for individuals in the dataset, not against the
model's internal feature representations.

Adding to this, no comparison with other theorem provers is provided in this work, as our main focus was to verify the ethical correctness of an AI system rather than the efficiency comparison of various theorem provers. Hence, this study focuses on the formalization and verification methodology and does not include empirical results, complexity analysis, or direct comparisons with alternative approaches. These aspects are left as future work to extend the framework towards practical evaluation and benchmarking. The implementation code and relevant datasets will be made available to ensure reproducibility.
\section{Conclusion}\label{conclusion}
This paper suggests the use of deontic logic along with temporal operators to formalize and evaluate ethical principles in AI. We provided fundamental properties that an ethical system should follow and developed theorems to validate its ethics in terms of the principles---fairness and explainability. We also observed experimentally the efficacy of this formalization in evaluating the ethical aspects of real-world AI systems.
The results demonstrated that the application of deontic logic and temporal operators to AI ethics represents a significant step forward in our ability to formally specify and verify the ethical behavior of AI systems. This work can help to identify potential ethical issues early in the development process and provide assurances that AI systems will behave following specified ethical principles. Moreover, the use of formal methods for AI ethics can facilitate the development of standardized approaches to ethical AI design and governance. By establishing a common language and framework for reasoning about AI ethics, this work can contribute to the creation of industry-wide standards and best practices. This, in turn, can help to build public trust in AI systems and ensure that the benefits of AI are realized responsibly and ethically. \par

While our approach offers a good starting point, future research will explore additional facets to increase its expressiveness and applicability. In future work, we plan to investigate normative ethics functionalities to create a framework capable of resolving ethical dilemmas and prioritizing actions at the individual level during conflicts. This is crucial for ensuring that ethical principles remain consistent, even when actions are complex or when there are competing priorities. While this paper focuses on verifying system-level ethical properties, further exploration into individual-level actions and their interactions within broader systems is necessary. 

Given the non-monotonic or evolving nature of AI, incorporating dynamic actions into this framework can lead to rapid growth in the knowledge base, especially when applied to large-scale models like Large Language Models (LLMs).To manage this challenge, we propose integrating visual logic techniques, such as constraint diagrams with temporal aspects~\cite{visual_logic_example}. These diagrams show promising potential in mitigating scalability issues that arise from purely logical approaches. By visualizing relationships between entities over time, these diagrams provide a flexible and intuitive method for expressing subset relationships and set cardinality constraints, ultimately aiding in clearer and more adaptable formalization. Overall, this foundational work aims to inspire and guide the development of scalable, adaptive, and practical frameworks for ethical AI systems. By tackling both theoretical and practical challenges, it seeks to pave the way for responsible and trustworthy AI deployment in various applications.

\balance
\bibliographystyle{IEEEtran}
\bibliography{bib}

\end{document}